\documentclass{article}
\usepackage{ijcai18}

\usepackage{times}
\usepackage{xcolor}
\usepackage{soul}
\usepackage[utf8]{inputenc}
\usepackage[small]{caption}

\usepackage{graphicx}

\usepackage{amssymb}
\usepackage{amsmath}

\DeclareMathOperator*{\argmax}{arg\,max}

\title{Improving Information Extraction from Images with Learned Semantic Models \thanks{This paper is an abridged version,  with some additional content, of a paper titled "Improving Visual Relationship Detection using Semantic Modeling of Scene Descriptions" published at the ISWC-2017.}}

\author{
Stephan Baier$^{1,2}$, 
Yunpu Ma$^{1,3}$, 
Volker Tresp$^{1,3}$
\\ 
$^1$ Ludwig Maximilian University Munich \\
$^2$ Data Reply GmbH Munich \\
$^3$ Siemens Corporate Technology Munich\\
stephan.baier@campus.lmu.de,
yunpu.ma@siemens.com,
volker.tresp@siemens.com
}

\begin{document}

\maketitle

\begin{abstract}
Many applications require an understanding of an image that goes beyond the simple detection and classification of its objects. In particular, a great deal  of semantic information is carried in the relationships between objects. We have previously shown that the combination of a visual model and a statistical semantic prior model can improve on the task of mapping images to their associated scene description. In this paper, we review the model and compare it to a novel conditional multi-way model for visual relationship detection, which does not include an explicitly trained visual prior model. We also discuss potential relationships between the proposed methods and memory models of the human brain.
\end{abstract}

\section{Introduction}

The extraction of semantic information from unstructured data is a key challenge in artificial intelligence. Object detection in images has improved enormously within the last years, due to novel deep learning methods. However, the semantic expressiveness of image descriptions that consist simply of a set of objects is rather limited.  Semantics is captured in more meaningful ways by the relationships between objects. In particular, visual relationships can be represented by triples, where two entities appearing in an image are linked through a relation (e.g. \textit{man-riding-elephant}, \textit{man-wearing-hat}). Due to the cubic combinatorial complexity of possible triples, it is likely that not all relevant triples do appear in the training data, which makes training a predictive model difficult. In this paper, we review our previously proposed approach published in \cite{baier_iswc}, which uses a Bayesian fusion approach for combining visual object detection methods with a separately trained probabilistic semantic prior. Incorporating a probabilistic semantic prior especially helps in cases where the prediction of the classifier is not very certain, and for the generalization to unobserved triples in the training set. Further, we propose a new conditional multi-way model which is inspired by statistical link prediction methods. This model does not include an explicitly trained prior of the semantic triples, and is trained in a purely feedforward manner. The prior is implicitly learned in the latent representations of the entities. We conduct experiments on the Stanford Visual Relationship dataset recently published by \cite{visual_relationship}. For the Bayesian fusion model we evaluate different model variants on the task of predicting semantic triples and the corresponding bounding boxes of the subject and object entities detected in the image. Our experiments show that including the semantic model improves on the state-of-the-art result in the task of mapping images to their associated triples. The experiments further show that the conditional multi-way model proposed in this paper, especially in the task of predicting unobserved triples, achieves performance that is comparable to the Bayesian fusion model.


\section{Background and Related Work}
\label{related_work}

In this section, we discuss the most important background and related work for visual relationship detection.

\subsection{Visual Relationship Detection}

Visual relationship detection is concerned with the problem of detecting objects and their relationships in images. Extracting triples, i.e. visual relationships, from raw images is a challenging task, which has been a focus in the Semantic Web community for some time, e.g. \cite{sw2,symbolic2} and recently also gained substantial attention in mainstream computer vision \cite{phrase1,visual_relationship,visual_transe}. In the approach from Lu et al. \cite{visual_relationship}, a Region Convolutional Neural Network (RCNN) for detecting and classifying objects in the image was used. Given a pair of objects, their relationship is predicted using another Convolutional Neural Network (CNN). The prediction is combined with a prior based on the word embeddings of the objects and the relationships. Baier et al. \cite{baier_iswc} have used the same RCNN, but have replaced the word embedding model with a
learned probabilistic semantic model. This model will be reviewed in Section \ref{fusion_model}. More recently, Zhang et al. \cite{visual_transe} have applied the  translational embedding model \cite{transe} for the task of visual relationship detection.

\begin{figure*}[t]
	\centering
	\includegraphics[width=0.82\textwidth]{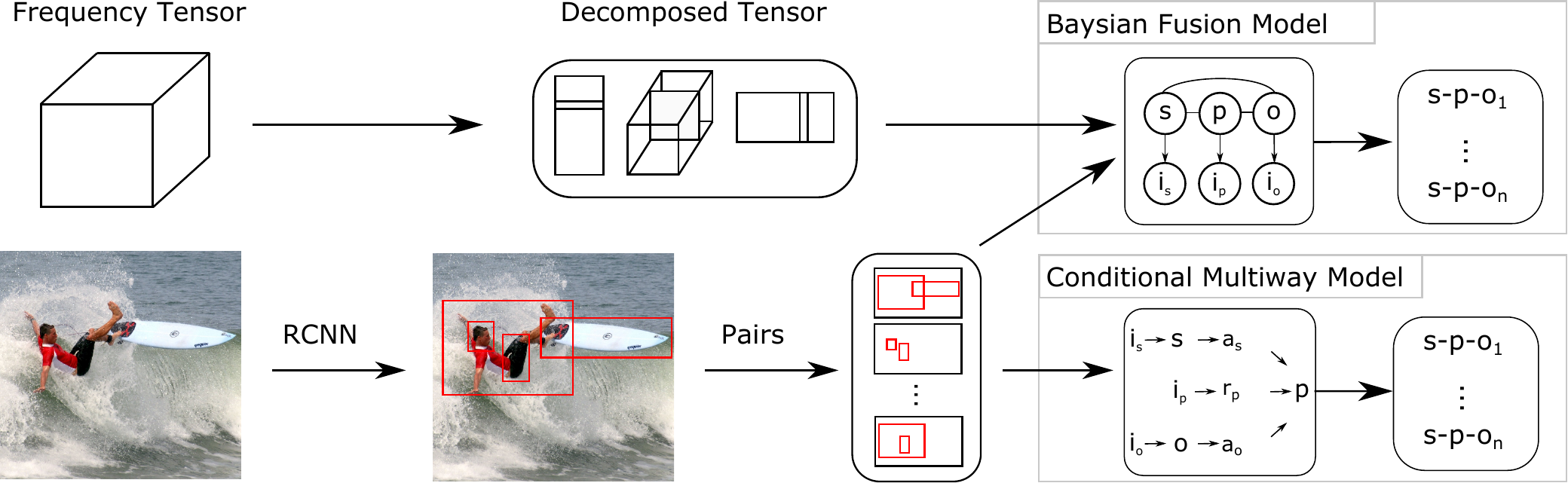}
	\caption{The procedure of deriving a list of triples given an image. The last step differs for the two proposed models.}
	\label{fig_pipeline}
\end{figure*}

\subsection{Semantic Tensor Models}
\label{semantic_tensor_models}

A number of statistical models have been proposed for modeling graph-structured knowledge bases, often referred to as knowledge graphs. A knowledge graph $\mathcal{G}$ consists of a set of triples $\mathcal{G} = \{(s, p, o)_i\}_{i=1}^N \subseteq \mathcal{E} \times \mathcal{R} \times \mathcal{E}$. The entities $s, o \in \mathcal{E}$ are referred to as \textit{subject} and \textit{object} of the triple, and the relation between the entities $p \in \mathcal{R}$ is referred to as \textit{predicate} of the triple.


Link prediction methods can be described by a function $\theta: \mathcal{E} \times \mathcal{R} \times \mathcal{E} \rightarrow \mathbb{R}$, which maps a triple $(s,p,o)$ to a real-valued score, which is a measure for the likelihood of the triple being true. Most recently developed link prediction models learn a latent representation, also called embedding, for the entities and the relations. In the following we describe the link prediction methods, which are used in this paper.

\paragraph{DistMult:} DistMult \cite{DistMult} scores a triple by building the tri-linear dot product of the embeddings, such that
\begin{equation}
\label{eq_link_first}
\theta(s, p, o) = \langle a_s, r_p, a_o\rangle
\end{equation}
where $a_s, r_p, a_o \in \mathbb{R}^d$ are latent vector representations for subject, predicate, and object, and  $\langle \cdot, \cdot, \cdot \rangle$ denotes the dot product of multiple vectors. The dimensionality $d$ of the embeddings, also called rank, is a hyperparameter of the model.

\paragraph{ComplEx:} ComplEx \cite{ComplEx} extends DistMult to complex-valued vectors for the embeddings of both, relations and entities. The score function is
\begin{equation}
\theta(s,p,o) = Re(\langle a_s, r_p, \overline{a_o}\rangle)
\end{equation}
where $a_s, r_p, a_o \in \mathbb{C}^d$ are complex-valued vector representations of subject, predicate and object. $Re(\cdot)$ denotes the real part of a complex number and $\overline{\cdot}$ denotes the complex conjugate.

\paragraph{Multiway NN:} The multiway neural network \cite{knowledge_vault,relational_review} concatenates all embeddings and feeds them to a neural network of the form
\begin{equation}
\theta(s,p,o) = \beta^T \tanh \left( W \left[a_s, r_p, a_o \right]  + b_1 \right) + b_2
\end{equation}
where $\left[\cdot, \cdot, \cdot \right] $ denotes the concatenation of the embeddings $a_s, r_p, a_o \in \mathbb{R}^d$. The prediction is derived using a Mulilayer Perceptron with the weight matrix $W \in \mathbb{R}^{3d \times z}$, the weight vector $\beta \in \mathbb{R}^{z}$, and the biases $b_1 \in \mathbb{R}^{z}, b_2 \in \mathbb{R}$.

\paragraph{RESCAL:} The tensor decomposition RESCAL \cite{Rescal} learns vector embeddings for entities and matrix embeddings for relations. The score function is
\begin{equation}
\label{eq_link_last}
\theta(s,p,o) = a_s \cdot R_p \cdot a_o
\end{equation}
with $\cdot$ denoting the dot product, $a_s, a_o \in \mathbb{R}^d$ and $R_p \in \mathbb{R}^{d \times d}$.


\subsection{Image Classification and Object Detection}

The Region Convolutional Neural Network (RCNN) \cite{rcnn} is a standard approach for detecting objects in images. It uses a selective search algorithm for getting candidate regions in an image. The RCNN algorithm then rejects most of the regions based on a classification score. As a result, a small set of region proposals is derived. Convolutional Neural Networks (CNNs) have become the standard approach for classifying images. CNNs apply convolutional filters in a hierarchical manner to an image. In this work, we use a specific CNN network architecture, which is called VGG-16 \cite{vgg}. It consists of 16 convolutional layers and two dense output layers. The output of the second last layer of the network can be considered as a latent representation of the input image.

\begin{table*}[h]
	\centering
	\setlength{\tabcolsep}{0.85em}
	{\renewcommand{\arraystretch}{1.4}
		\begin{tabular}{|c||c|c||c|c||c|c||c|c|}
			\hline
			Task &\multicolumn{2}{c||}{Phrase Det.} &\multicolumn{2}{c||}{Rel. Det.} &\multicolumn{2}{c||}{Predicate Det.} & \multicolumn{2}{c|}{Triple Det.} \\
			\hline Evaluation & R@100 & R@50 & R@100 & R@50 & R@100 & R@50 & R@100 & R@50\\
			\hline
			\hline Lu et al.  V \cite{visual_relationship} & 2.61 & 2.24 & 1.85 & 1.58 & 7.11 & 7.11 & 2.68 & 2.30 \\ 
			\hline Lu et al. full \cite{visual_relationship} & 17.03 & 16.17 & 14.70 & 13.86 & 47.87 & 47.87 & 18.11 &  17.11 \\ 
			\hline \hline Conditional Multiway Model & 17.71 & 15.79 & 15.37 & 13.72 & 47.93 & 47.62 & 18.53 & 16.47  \\ 
			\hline \hline  RESCAL Prior & 19.17  & 18.16  &  16.88  & 15.88  & 52.71  & 52.71 & \textbf{20.23} & \textbf{19.13}  \\
			\hline   MultiwayNN Prior & 18.88 & 17.75 & 16.65 & 15.57 & 51.82 & 51.82 & 19.76 & 18.53 \\
			\hline   ComplEx Prior & \textbf{19.36} & \textbf{18.25} & \textbf{17.12} & \textbf{16.03} & \textbf{53.14} & \textbf{53.14} & \textbf{20.23} & 19.06 \\ 
			\hline  DistMult Prior & 15.42 & 14.27 & 13.64 & 12.54 & 42.18 & 42.18 & 16.14 & 14.94 \\
			\hline 
		\end{tabular} 
	}
	\caption{Results for visual relationship detection. We report Recall at 50 and 100 for four different validation settings.}
	\label{results_main}
\end{table*}

\section{Modelling Visual Relationships}

In this section, we present two different models for the task of visual relationship detection, both combining semantic tensor models and object detection in different ways. Figure \ref{fig_pipeline} shows the processing pipeline for both models. Both assume that object candidate boxes are provided by an RCNN model. The goal is to predict the most likely triple $(s,p,o)$ for each pair of subject/object candidate boxes $(i_s, i_o)$. We define the union of the regions $i_s$ and $i_o$ as $i_p$. The extracted triples then consist of two visual concepts $s,o$ and their relationship $p$. This is different to knowledge graphs where the relations are typically not modelled on the concept level, but on the instance level. Nevertheless,  the link prediction methods described in Section \ref{semantic_tensor_models} can be applied to  visual concepts, as well.
\subsection{Bayesian Fusion Model}
\label{fusion_model}

In this model, we derive predictions from two different CNNs, one modelling $p(s|i_s)$ and $p(o|i_o)$, and the other one modelling $p(p|i_p)$. We combine these visual models with a tensor model serving as a semantic prior $p(s,p,o)$ in a Bayesian way.
We assume the joint distribution of all involved variables to factor as
\begin{equation}
p(s, p, o, i_s, i_p, i_o) \propto \tilde{p}(s,p,o) \cdot \tilde{p}(i_s|s) \cdot \tilde{p}(i_p | p) \cdot \tilde{p}(i_o | o)
\label{eq_joint_prob}
\end{equation}
with $\tilde{p}$ denoting unnormalized probabilities. We can divide the joint probability of Equation \ref{eq_joint_prob} into two parts. The first part is $\tilde{p}(s,p,o)$, which models semantic triples. The second part is $\tilde{p}(i_s|s) \cdot \tilde{p}(i_p | p) \cdot \tilde{p}(i_o | o)$, which models the visual part given the semantics. 
The semantic prior is modelled as
\begin{equation}
\tilde{p}(s, p, o) = \exp(\theta(s,p,o)),
\end{equation}
where $\theta$ is a semantic tensor model as described in Section \ref{semantic_tensor_models} and $\exp$ is the activation function for Poisson regression.  An advantage is that the semantic model can be trained separately from the visual model using only the absolute frequencies of triples in the training data. As we are predicting count data, we train the model using a Poisson cost function.\footnote{Another sampling model would lead to  a multinomial model, which would only result in  a different normalization of the distribution.}
The semantic model and the visual models are combined by applying Bayes rule to Equation \ref{eq_joint_prob}, such that
\begin{equation}
p(s, p, o| i_s, i_p, i_o) \propto \tilde{p}(s,p,o) \cdot \frac{\tilde{p}(s | i_s) \cdot \tilde{p}(p | i_p) \cdot \tilde{p}(o | i_o) }{\tilde{p}(s) \cdot \tilde{p}(p) \cdot \tilde{p}(o)}.
\end{equation}
The additional terms of the denominator $\tilde{p}(s)$, $\tilde{p}(p)$, $\tilde{p}(o)$ are derived through the marginalization of $\tilde{p}(s, p, o)$ and a Laplacian smoothing. For each pair of bounding boxes, we pick the triple with the highest probability.

\subsection{Conditional Multiway Model}

In this model, we derive for each pair of bounding boxes the subject $s$ and the object $o$ by applying a VGG classifier to the regions $i_s$ and $i_o$. In a second step we then build a conditional tensor model for $p(p|s,o,i_p)$. We parameterize the model as
\begin{equation}
p(p|s,o,i_p) = \text{softmax} (W_2 \tanh(W_1 \left[ a_s, a_o, r_{i_p} \right] + b_1) + b_2),
\label{eq:forward}
\end{equation}
with $a_s, a_o \in \mathbb{R}^d$ being latent vector representations for the visual concepts, $r_{i_p} \in \mathbb{R}^d$ being a latent representation vector for the image patch $i_p$, and $\left[\cdot, \cdot, \cdot \right] $ denoting the concatenation operation. For $a_s$ and $a_o$ the representations are optimized in the learning procedure and stored in a lookup table. For deriving a representation of the predicate region $i_p$, we model $r_{i_p} = M h_{i_p}$, where $h_{i_p}$ is the activation of the second last layer of a VGG network with the image region $i_p$ as input. The matrix $M$ maps the latent representation of the VGG network to a vector with the rank of the multi-way model. The probabilities for predicate $p$ are derived by applying a Multilayer Perceptron with the additional parameters $W_1 \in \mathbb{R}^{3d\times z}, W_2 \in \mathbb{R}^{z\times |\mathcal{R}|}, b_1 \in \mathbb{R}^z, b_2 \in \mathbb{R}^{|\mathcal{R}|}$. 

To derive a single prediction for each pair of bounding boxs, we pick the subject $\hat{s} = \argmax p(s| i_s)$, object $\hat{o} = \argmax p(o|i_o)$, and predicate $\hat{p} = \argmax p(p | i_p, \hat{s}, \hat{o})$ with the highest probabilities. The confidence score for the triple $(\hat{s}, \hat{p}, \hat{o})$, given an input region $(i_s, i_p, i_o)$ is calculated as
\begin{equation}
p(\hat{s}, \hat{p}, \hat{o} | i_s, i_p, i_o) = p(\hat{s} | i_s) \cdot p(\hat{o} | i_o) \cdot  p(\hat{p} | i_p, \hat{s}, \hat{o}).
\end{equation}


\begin{table*}[h]
	\centering
	\setlength{\tabcolsep}{0.85em}
	{\renewcommand{\arraystretch}{1.4}
		\begin{tabular}{|c||c|c||c|c||c|c||c|c|}
			\hline
			Task &\multicolumn{2}{c||}{Phrase Det.} &\multicolumn{2}{c||}{Rel. Det.} &\multicolumn{2}{c||}{Predicate Det.} & \multicolumn{2}{c|}{Triple Det.} \\
			\hline Evaluation & R@100 & R@50 & R@100 & R@50 & R@100 & R@50 & R@100 & R@50\\
			\hline
			\hline Lu et al. V \cite{visual_relationship} & 1.12 & 0.95 & 0.78 & 0.67 & 3.52 & 3.52 & 1.20 & 1.03 \\ 
			\hline Lu et al. full \cite{visual_relationship} & 3.75 & 3.36 & 3.52 & 3.13 & 8.45 & 8.45 & 5.39 & 4.79 \\ 
			\hline \hline Conditional Multiway Model & 5.73 & 5.39 & 5.22 & 4.96 & 14.32 & 14.32 & 5.22 & 4.96 \\
			\hline \hline RESCAL Prior & 6.59 & \textbf{5.82} & 6.07 & \textbf{5.30} & 16.34 & 16.34 & 6.07 & \textbf{5.30}\\
			\hline  MultiwayNN Prior & \textbf{6.93} & 5.73 & \textbf{6.24} & 5.22 & \textbf{16.60} & \textbf{16.60} & \textbf{6.24} & 5.22 \\
			\hline  ComplEx Prior & 6.50  & 5.73 & 5.82 & 5.05 & 15.74 & 15.74 & 5.82 & 5.05 \\
			\hline  DistMult Prior & 4.19 & 3.34 & 3.85 & 3.08 & 12.40 & 12.40 & 3.85 & 3.08 \\
			\hline 
		\end{tabular} 
	}
	\caption{Results for the zero shot learning experiments. We report Recall at 50 and 100 for four different validation settings.}
	\label{results_zeroshot}
\end{table*}

\section{Experiments}


We evaluate our proposed method on the recently published Stanford Visual
Relationship dataset. 

\paragraph{Setting}
We compare the two models presented in this paper and their variations, with the results from \cite{visual_relationship}. The settings are the same as in \cite{visual_relationship} and \cite{baier_iswc}. In all settings a single triple is derived for each pair of bounding boxes. In the first setting, which in \cite{visual_relationship} is referred to as \textit{Phrase Detection}, a triple with its corresponding bounding boxes is considered correctly detected, if the triple is similar to the ground truth, and if the union of the bounding boxes has at least 50 percent overlap with the union of the ground truth bounding boxes. In \textit{Relationship Detection}, both the bounding box of the subject and the bounding box of the object need at least 50 percent of overlap with their ground truth. In \textit{Predicate Detection},  it is assumed that subject and object are given, and only the correct predicate linking both needs to be predicted. In \textit{Triple Detection}, a triple is considered correct if it corresponds to the ground truth, independent of the predicted bounding boxes. 


\paragraph{Results}
Table \ref{results_main} shows the results for visual relationship detection. The first row shows the results, when only the visual part of the model is applied. This model performs poorly in all four settings. The full model in the second row adds the language prior to it, which drastically improves the results. The \textit{Conditional Multiway Model} outperforms the language prior model in some settings and achieves very similar results in the others. In the last four rows we report the results of the Bayesian fusion model, with different link prediction methods. We see that the model performs consistently better than the state-of-the-art method proposed by \cite{visual_relationship}. Only \textit{DistMult} is slightly worse, which might be due to the fact that it assumes symmetric scores when subject and object are exchanged.

Table \ref{results_zeroshot} shows the results, when only evaluating triples which have not been observed in the training data. This task is much more difficult, as it requires the models to generalize to these triples. Also in this experiment, including the semantic model significantly improves the prediction. For the first three settings, the best performing method, which is the \textit{Multiway Neural Network}, almost retrieves twice as many correct triples, as the state-of-the-art model of \cite{visual_relationship}. These results clearly show that our model is able to infer also new likely triples, which have not been observed in the training data. The \textit{Conditional Multiway Model} achieves a performance close to the Bayesian fusion models, although it does not include a separately learned prior for the semantic triples.

\section{Relationships to Perception and Memory}

At a higher cognitive level, information which is ceaselessly acquired by the visual system of the brain, needs to be interpreted correctly. Goethe's proverb might be quite fitting for our proposed approach: ''You only see what you know'' since one can only perceive things for which we have an internal brain  representation. This internal representation  is used by the perceptual system and by the main declarative memory systems, i.e., the episodic memory (about events we \textit{remember}) and the semantic memory system (about facts we \textit{know})~\cite{brain_tulving}. Perception even needs more: it needs to generalize to novel entities on the class level, since a subject constantly encounters novel entities: The perceptual system needs to be able  to  generalize to new scenes and needs to generalize on the associations of new perceptual components.  According to the complementary learning system \cite{brain_mcclelland}, the memory systems are the basis for the brain to learn to generalize to new situations, a  consolidation process which might happen largely during sleep and might involve the neocortex. Our work focuses on the generalization after training the perceptual system for scene comprehension. Our  approaches can be related to some of the main current hypothesis about perception and memory. Many groups favor the Bayesian brain hypothesis, which assumes that the brain uses inherited and learned prior hypothesis to understand and reason about the world (\cite{knill} and \cite{griffiths} are two examples). Our Bayesian fusion model fits precisely into this category. Its tensor model provides a  prior model which is quite powerful in supporting the perceptual pipeline. It is a very rich prior compared to simple smoothness priors used in other approaches. In contrast, in the  conditional multiway model the prior is represented in the latent representations of entities and weights in the feedforward neural network. The conceptual memory is formed implicitly in the end-to-end training of the model. Cognitive models which are pursuing this approach have been proposed and studied in \cite{tresp1,tresp2,tresp3}. This approach is more in favor of a theory which assumes that the brain, at least conceptionally, is trained end-to-end with few clearly interpretable functional modules.

\section{Conclusion}

We presented two approaches for visual relationship detection, which both include statistical semantic models. The first approach, which originally has been published in \cite{baier_iswc}, combines standard computer vision methods with latent variable models for link prediction. We proposed a probabilistic framework,  in form of a Bayesian fusion model,  for integrating both the semantic prior and the computer vision algorithms into a joint model. The second approach uses a conditional multi-way model, which is inspired by link prediction methods. For the prediction of triples, which have not been observed in the training data, the performance of the second approach is on par with the first approach, as its structure helps to generalize to unobserved triples, without including a separately trained prior for the semantic triples. Both approaches form statistical models on the class level, and can thus generalize to new images. This is in contrast to typical knowledge graph models, where nodes correspond to specific instances. In cognitive terms, the  Bayesian fusion model can more directly be related to the Bayesian brain hypothesis, as being pursued by many research teams, whereas the conditional multiway model is more closely related to the tensor memory hypothesis \cite{tresp3}.

\bibliographystyle{named}
\bibliography{ijcai18}

\end{document}